\documentclass[10pt,journal,cspaper,compsoc]{IEEEtran}
\ifCLASSOPTIONcompsoc
\else
\fi
\ifCLASSINFOpdf
   \usepackage[pdftex]{graphicx}
\else
\fi
\usepackage[noend]{algorithmic}
\usepackage{boxedminipage}

\begin{document}
%
\title{Technical Report CMPC14-03: Finding Motif Sets in Time Series}
%
%
%
%

\author{Anthony~Bagnall, ~Jon~Hills
        and~Jason~Lines,
\IEEEcompsocitemizethanks{\IEEEcompsocthanksitem J. Hills, J. Lines, and A.Bagnall are with the School
of Computing Sciences, University of East Anglia, Norwich, Norfolk, United Kingdom.\protect\\
E-mail: \{ajb,j.hills,j.lines\}@uea.ac.uk}
\thanks{}}

%
%

\markboth{University of East Anglia Computer Science Technical Report CMPC14-03}%
{Shell \MakeLowercase{\textit{et al.}}: Bare Demo of IEEEtran.cls for Computer Society Journals}
%


\IEEEcompsoctitleabstractindextext{%
\begin{abstract}
Time-series motifs are representative subsequences that occur frequently in a time series; a motif set is the set of subsequences deemed to be instances of a given motif. We focus on finding motif sets. Our motivation is to detect motif sets in household electricity-usage profiles, representing repeated patterns of household usage.

We propose three algorithms for finding motif sets. Two are greedy algorithms based on pairwise comparison, and the third uses a heuristic measure of set quality to find the motif set directly. We compare these algorithms on simulated datasets and on electricity-usage data. We show that Scan MK, the simplest way of using the best-matching pair to find motif sets, is less accurate on our synthetic data than Set Finder and Cluster MK, although the latter is very sensitive to parameter settings. We qualitatively analyse the outputs for the electricity-usage data and demonstrate that both Scan MK and Set Finder can discover useful motif sets in such data.
\end{abstract}

}

\maketitle

\IEEEdisplaynotcompsoctitleabstractindextext

%
\IEEEpeerreviewmaketitle

\section{Introduction}
%
%

%
%
%
%
\IEEEPARstart{T}{ime-series} motifs are subsequences that occur frequently in a time series \cite{lonardi2002finding}. They can be used to characterise the typical behaviour of a time series to allow for classification or anomaly detection (e.g. \cite{lonardi2002finding}), or as a primitive in, for example, association rule mining (e.g. \cite{das1998rule}). Areas of application include medicine~\cite{lonardi2002finding}, image processing~\cite{lonardi2002finding}, and robotics~\cite{oates2000method}. Our interest lies in finding motifs in household electricity-usage profiles, and using them to disaggregate the data in terms of devices~\cite{froehlich2011disaggregated}.

The key contributions to the study of motifs are presented in \cite{lonardi2002finding,mueen2009exact,mueen2011disk}. The algorithm described in \cite{lonardi2002finding} operates on discretised time series. This approximate approach is inappropriate for our problem, as the electricity-usage data is already aggregated into 15-minute periods, a time frame that makes device disambiguation difficult. Further compression would make detection impossible. The emphasis of the later works is on finding best-matching pairs of subsequences. Our contribution is directed at exact discovery of frequently occurring subsequences, rather than best-matching pairs. We propose three algorithms for this purpose. Two (Sections \ref{iterative} and \ref{clustering}) use the MK pair-matching algorithm as a subroutine (although any pair-finding algorithm could be used), and the other finds motif sets based on whole set quality, rather than pair matching (Section~\ref{brute}). We assess their performance (Sections~\ref{results} and \ref{elecRes}) on both real and synthetic data, described in Section~\ref{data}. All of the code and data used in this paper can be found at~\cite{icdmWeb}; the password for the zipped files is `motset13'.

\begin{figure}[!t]
\centering
\includegraphics[width=3in,height=1in]{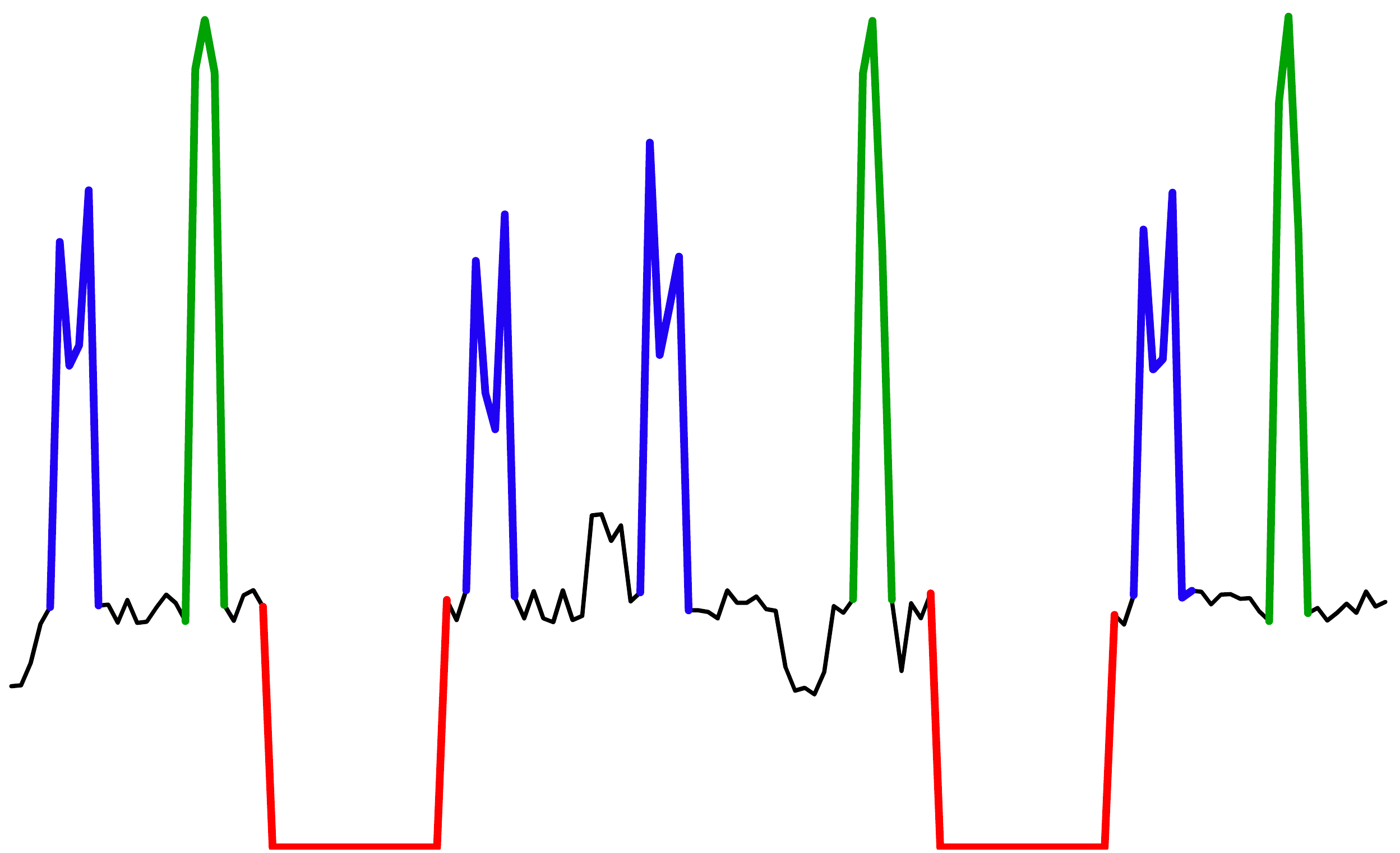}
\caption{A simulation of an electricity demand profile, with three motif sets.}
\label{exProfile}
\end{figure}

\section{Background}
\label{background}

A \emph{time series} is a sequence $T = <t_1, t_2,...,t_m>$ of $m$ real-valued numbers. Images and spectrographs \cite{bagnall12ensembles}, and other sequential data, as well as temporally-ordered data, may be regarded as time series. Time-series data-mining research is concerned predominantly with discovering similarities between subsequences of time series, rather than between whole time series, see e.g. \cite{lonardi2002finding}. A \emph{subsequence} of length $n$ of a time series $T$ is a time series $S_{a} = <t_a, t_{a+1},...,t_{a+n-1}>$ where $1 \leq a \leq m - (n - 1)$. A \emph{sliding window} produces the set $S$ of all possible subsequences $S_{i}$ of size $n$ of $T$. The cardinality of $S$ is $m-(n-1)$.

We use the term {\em motif} to refer to a single subsequence (which can be a concrete instance, or the average of the members of its motif set), and the term {\em motif set} to mean the set of subsequences that are associated with a given motif. The 1-motif set problem is to find the largest subset of $S$ whose members are considered to {\em match} one another, where two series match if the distance between them is less than some threshold parameter, $r$, and the match is {\em non-trivial}.
As shown in \cite{keogh2005clustering}, failing to prevent trivial matching renders motif detection meaningless. There are alternative definitions of a trivial match; we adopt the definition used in \cite{keogh2005clustering}, and take it that two series cannot match if they overlap.

The $K$-motif set is defined as the $K^{th}$ most commonly-occurring subsequence that does not overlap with the previous $(K-1)$-motif sets. Finding $K$ motif sets requires that we enforce a separation of at least $2r$ between each motif set and all previous motif sets~\cite{lonardi2002finding}.

\subsection{Mueen-Keogh (MK) Best-matching Pair Algorithm.}

An exact pair-finding algorithm called Mueen-Keogh (MK) is proposed in \cite{mueen2009exact}. MK finds closest matches between time-series subsequences using a form of early abandon that dramatically speeds up the matching process in the average case. Finding best-matching pairs is of less interest to us than detecting the daily repeating pattern of, for example, a washing machine or oven. This is an important point because the best-matching pair are not necessarily members of a high-cardinality motif set; for example, in Fig.~\ref{exProfile}, the red subsequences form the best-matching pair, but the blue and green motif sets have higher cardinality. A formal description of MK is given in~\cite{mueen2009exact}; we use the algorithm to find best-matching pairs as one stage in the motif-set discovery process; any pair-finding algorithm can be substituted for MK, however.




\section{Finding the $K$-Motif Sets}
\label{motifSet}

In this section, we describe three algorithms for finding the $K$-motif sets. Two are based on constructing sets using pairs, the other constructs motif sets directly. MK is a fast way of finding matching pairs; however, this does not in itself provide a way of finding motif sets.

\subsection{Scan MK.}
\label{iterative}

We have extended the method for finding the range motif outlined in~\cite{mueen2011disk} to find approximate $K$-motif sets. We iterate the process of finding closest pairs and their matches, adding them to a motif set and removing members and their trivial matches from the list of candidates after each iteration. The algorithm is described in Fig.~\ref{iterativeMK}. We assume a distance function $d(S_i,S_j)$ is defined (we use Euclidean distance for all experiments).

MK is used to find the best-matching pair of subsequences in $S$ (line 7); if the distance between them is greater than $2r$, the algorithm terminates. Otherwise, the best-matching pair is added to a motif set, the trivial matches of the best-matching pair are removed from $S$ (lines 8-16), and the remaining subsequences are scanned. Any subsequences within $2r$ of both members of the best-matching pair are added to the motif set (lines 17-22).

The \texttt{condense} function operates as follows. For each set of contiguously-indexed subsequences, the non-trivial match is taken to be the subsequence with the smallest total distance between it and each of the members of the motif set. The contiguously-indexed subsequences are taken to be trivial matches, and are excluded from the motif set. The motif set is further condensed by removing one subsequence of any pair whose members are more than $2r$ apart. We choose which subsequence to exclude based on the number of clashes. For example, if a subsequence that clashes with three others is greater than $2r$ from a subsequence that clashes with two others, the first subsequence is removed, maximising the cardinality of the set. Ties are decided based on average linkage; the subsequence with the shortest total distance to the other members of the set is retained. Once the motif set is established, its members and their trivial matches are removed from the candidate set, and the process is repeated until no more subsequences are within $2r$ of each other.

\renewcommand{\algorithmicrequire}{\textbf{Input:}}
\renewcommand{\algorithmicensure}{\textbf{Output:}}

\begin{figure}[htp]
\begin{boxedminipage}{3.4in}
\caption{Scan MK}
\label{iterativeMK}
\begin{algorithmic}[1]
\REQUIRE $T$: a time series.\\
 $q$: the number of subsequences for MK.\\
 $r$: half the maximum width of a motif set.\\
 $n$: the width of the sliding window.\\
\ENSURE $M$: a set of motif sets.\\
\STATE $M \leftarrow \emptyset$
\STATE $F \leftarrow  \mathtt{slidingWindow}(T,n)$\\
\COMMENT{{\em F is the full set of subsequences of length n}}
\STATE $S \leftarrow  F$
\STATE $k \leftarrow 0$
\WHILE{$end$ = \textbf{false}}
\STATE $end \leftarrow$ \textbf{true}
\STATE $\{L_1,L_2\} \leftarrow \mathtt{MK}(S, q)$
\COMMENT{{\em $L_1$ and $L_2$ are indexes in $F$}}
\IF{$d(F_{L_1},F_{L_2})\leq 2r$}
\STATE$end \leftarrow$ \textbf{false}
\STATE $k\leftarrow k+1$
\STATE $M_{k} \leftarrow \{F_{L_1},F_{L_2}\}$
\STATE $D \leftarrow \emptyset$
\FOR{$i\leftarrow 1$ \textbf{to} $|S|$}
\IF{$\mathtt{trivialMatch}(F_{L_1},S_{i})$\\$ \vee \; \mathtt{trivialMatch}(F_{L_2},S_{i})$}
\STATE $D \leftarrow D \cup S_{i}$
\ENDIF
\ENDFOR
\STATE $S \leftarrow S - D$
\FOR{$i\leftarrow 1$ \textbf{to} $|S|$}
\IF{$d(F_{L_1},S_{i})\leq 2r \wedge d(F_{L_2},S_{i})\leq 2r \wedge S_{i} \notin D$}
\STATE $M_{k} \leftarrow M_{k} \cup S_{i}$
\FOR{$j\leftarrow 1$ \textbf{to} $|S|$}
\IF{$\mathtt{trivialMatch}(S_{i},S_{j})$}
\STATE $D \leftarrow D \cup S_{j}$
\ENDIF
\ENDFOR
\ENDIF
\ENDFOR
\STATE $S \leftarrow S - D$
\STATE $M_k \leftarrow \mathtt{condense}(M_k,r$)
\ENDIF
\ENDWHILE
\IF{$k > 0$}
\STATE $M \leftarrow \{M_1,...,M_{k}\}$
\STATE $\mathtt{sort}(M)$
\ENDIF
\RETURN $M$
\end{algorithmic}
\end{boxedminipage}
\end{figure}

\subsection{Cluster MK.}
\label{clustering}
The second algorithm we develop is based on hierarchical clustering of best-matching pairs. Hierarchical clustering is a widely used clustering approach, see for example \cite{keogh2005clustering}, based on finding best-matching pairs of series. We use MK to find the pairs, and an adapted form of bottom-up hierarchical clustering described in Fig.~\ref{cluster}.

\begin{figure}[htp]
\begin{boxedminipage}{3.4in}
\caption{Cluster MK}
\label{cluster}
\begin{algorithmic}[1]
\REQUIRE $T$: a time series.\\
 $q$: the number of subsequences for MK.\\
 $r$: the radius of the motif clusters.\\
  $n$: the width of the sliding window.\\
\ENSURE $S$: a set of motifs.
\STATE $F \leftarrow  \mathtt{slidingWindow}(T,n)$
\STATE $S \leftarrow  F$
\WHILE{$end$ = \textbf{false}}
\STATE $end \leftarrow$ \textbf{true}
\STATE $\{L_1,L_2\} \leftarrow \mathtt{MK}(S, q$)
\IF{$\mathtt{d}(F_{L_1},F_{L_2})\leq r$}
\STATE $end \leftarrow$ \textbf{false}
\STATE $S\leftarrow S - \{F_{L_1},F_{L_2}\}$
\STATE $c \leftarrow merge(F_{L_1},F_{L_2})$
\STATE $S\leftarrow S \cup c$
\ENDIF
\ENDWHILE
\RETURN $S$
\end{algorithmic}
\end{boxedminipage}
\end{figure}

We find the closest pair of subsequences, then merge this pair to form a new cluster (motif set). The cluster is represented by a new subsequence found by averaging the input subsequences, weighted by the number of subsequences that have already been combined to make each candidate. This ensures the cluster centre accurately reflects the members of the cluster. The process is repeated until the distance between the best-matching pair is greater than $r$. At this point the subsequence set $S$ will contain the motifs, and the motif sets can be recovered from the clustering data structure.

\subsection{Set Finder.}
\label{brute}
We propose an algorithm to find the $K$-motif sets directly, based on counting and separating (Fig.~\ref{bruteAl}). Each subsequence is compared to every other subsequence, and the non-trivial matches are counted. The set of counts is sorted. The sorted set is then input to the function \texttt{separate}, which checks each subsequence with a non-zero count in order to ensure that it is at least $2r$ apart from subsequences with a greater number of matches. Subsequences that fail the test are removed from the set. An early abandon based on the value of $r$ is built into the distance function to speed up the algorithm.

\begin{figure}[htp]
\begin{boxedminipage}{3.4in}
\caption{Set Finder}
\label{bruteAl}
\begin{algorithmic}[1]
\REQUIRE $T$: a time series. \\
$r$: the maximum distance between matches.\\
$n$: the width of the sliding window.
\ENSURE $M$: a set of motif sets.
\STATE $S \leftarrow  \mathtt{slidingWindow}(T,n)$
\STATE $C \leftarrow <0,\ldots,0>$\\
\COMMENT{{\em $C$ is counts vector of length $|S|$ initialised to 0}}
\FOR{$i\leftarrow 1$ \textbf{to}  $|S|$}
\FOR{$j \leftarrow i+1$ \textbf{to} $|S|$}
\IF{$\mathtt{d}(S_{i},S_j) \leq r \; \wedge$\\
 $ \mathtt{trivialMatch}(S_{i},S_j) $= \textbf{false}}
\STATE $C_i \leftarrow C_i+1$
\STATE $C_j \leftarrow C_j+1$
\ENDIF
\ENDFOR
\ENDFOR
\STATE $\mathtt{sort}(C,S)$
\STATE$M \leftarrow \mathtt{separate}(C,S)$
\RETURN $M$
\end{algorithmic}
\end{boxedminipage}
\end{figure}

The storing and recovery of the motif sets is omitted for clarity, but is easily achieved by retaining references to subsequences in addition to count data (see \cite{icdmWeb}).

\section{Data}
\label{data}
Section~\ref{synth} describes the synthetic datasets we generate in order to test for statistically significant differences between the algorithms. Section~\ref{elec} describes the electricity-usage profiles we mine for motif sets.

\subsection{Synthetic Data.}
\label{synth}

We specify a parameterised data space from which datasets are drawn, and randomly generate independent datasets for a given set of parameters. The simulated data is white noise (observations of i.i.d. normally-distributed random variables with $\mu=0$ and $\sigma=1$) with shapes added to the noise at random intervals (see \cite{icdmWeb} for more details).



\begin{figure}[!t]
\centering
\includegraphics[width=3.2in]{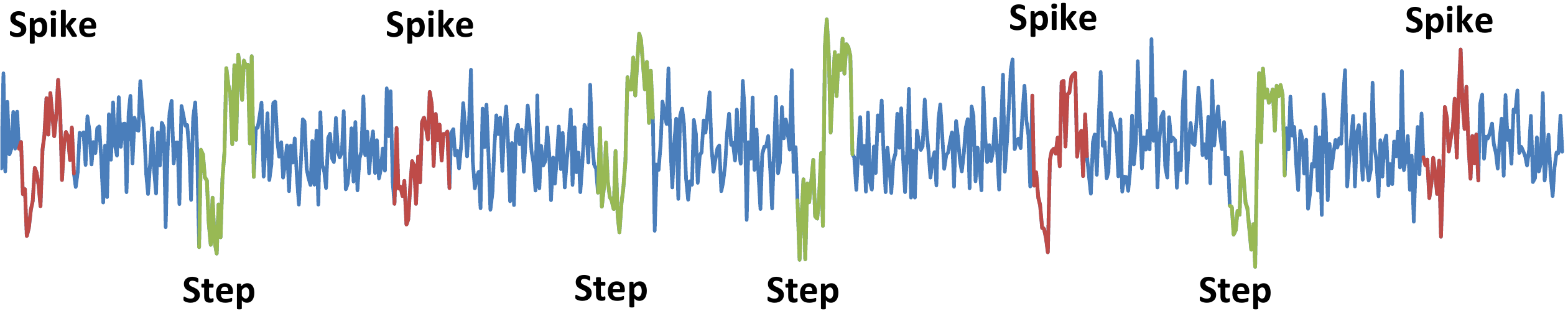}
\caption{An example simulated motif problem, with two motif sets, a Spike set (highlighted in red), and a Step set (highlighted in green).}
\label{fig_4}
\end{figure}

The minimum and maximum time series length, number of distinct shapes, and instances of each shape, are fixed parameters of the data, as are the length and amplitude of each instance. To generate a dataset, we randomly select one or two different shapes, and a number of instances for each shape. The shapes are added to the white noise at random locations, and do not overlap. An example of this distorted motif data is shown in Figure~\ref{fig_4}.

\subsection{Electricity-usage Data.}
\label{elec}
The electrical device data originate from a trial of smart meters in 187 homes across the United Kingdom (see~\cite{lines11,bagnall12ensembles}). The equipment was used to monitor the electricity consumption of each household in Watt Hours (Wh) at 15-minute intervals. Each series corresponds to the entire consumption of a household over the duration of the trial. The motivation for using this data is that the UK government has mandated that all households must be equipped with smart metering equipment by 2020. As a consequence, there will be very large quantities of data that must be processed in an efficient manner. 



One confounding factor is that devices of a similar nature have very similar usage profiles. Devices such as fridges and freezers, or computers and televisions, are very difficult to distinguish. In addition, the device-specific data is user-orientated. There is no central control over the devices that are monitored; the consumers have direct access to the monitoring equipment and all device labels are user-specified. Hence, labelling is potentially unreliable. Because of these confounding factors, it would be beneficial to have a reliable, automated method of detecting and identifying specific device use.

\section{Synthetic Data Results}
\label{results}



Our primary aim is to assess how accurately the algorithms discover motif sets; we include timing results on the companion website \cite{icdmWeb} for completeness. Cluster MK is slower than the other two algorithms on the synthetic data; Set Finder is slower on the electricity data, though the difference is marginal.

\subsection{Performance Evaluation.}
\label{perform}

For synthetic data containing only one motif set, we assess performance by calculating the number of \emph{true positives} (TP), \emph{false positives} (FP), and  \emph{false negatives} (FN). A TP occurs when the algorithm returns an index within $\frac{n}{2}$ of any shape, where $n$ is the length of the shape (in this case, fixed at 29). Any index returned by the algorithm that is not a TP is an FP, and any shape in the data that is not associated with a TP is an FN. We use two standard measures of accuracy: $precision = TP/(TP+FP)$, and $sensitivity = TP/(TP+FN)$.


For synthetic data containing two motif sets, the sets of indexes generated by the algorithm are paired with the indexes of the motif sets in the data, giving a combination we refer to as a \emph{matching}. A score is calculated for each matching as follows. Each index that is not paired with another index adds the value of $n$ (the length of the shape in the data) to the score. In our experiments, $n$ was fixed at 29. For each pair of indexes, the absolute value of the difference between the two is calculated, with a ceiling fixed at the value of $n$. Hence, pairing an index of 13 with an index of 21 gives a score of 8. The scores are tallied for each possible combination of indexes within sets, and of matchings between sets. Our measure rewards close matches, and punishes false negatives and false positives equally.

To assess significance for differences between algorithms at a given value of $r$, we perform a two-sample T-test with an alpha value of 0.05.



We compare the algorithms on two problems: finding a single set of shapes inserted into random noise, and finding two sets of shapes inserted into random noise. The second problem is more complex, and more representative of real-life applications. We use a range of $r$ values for two reasons. First, we are interested in discovering the value of $r$ that is appropriate for various situations. Second, we are interested in how the algorithms compare to one another over a range of values; if we used a single value for $r$, our results might be misleading.

\subsection{$r$ value.}

For the synthetic data, all algorithms perform best when $r$ is around $\frac{n}{2}$ for motif length $n$. We speculate that the high level of noise in the data prevents successful discovery of motifs with smaller values of $r$. For noisy data, we recommend setting $r$ at this level as a heuristic. If $r$ is higher than $\frac{n}{2}$ on our synthetic data, performance begins to degrade.

The electricity data is much less noisy than the synthetic data. On that data, we achieve good results with $r = \frac{n}{8}$. This equates to an $r$ value of 3.6 for the synthetic data, a value at which the sensitivity is 0. The synthetic data is too noisy to tolerate such a low value of $r$. As a general rule, we suggest indexing $r$ to $n$, and decreasing the value of $r$ as the level of noise in the data decreases.

\subsection{Problems with a single motif set.}


\begin{figure}[!t]
\centering
\includegraphics[width=1.6in,height=2in]{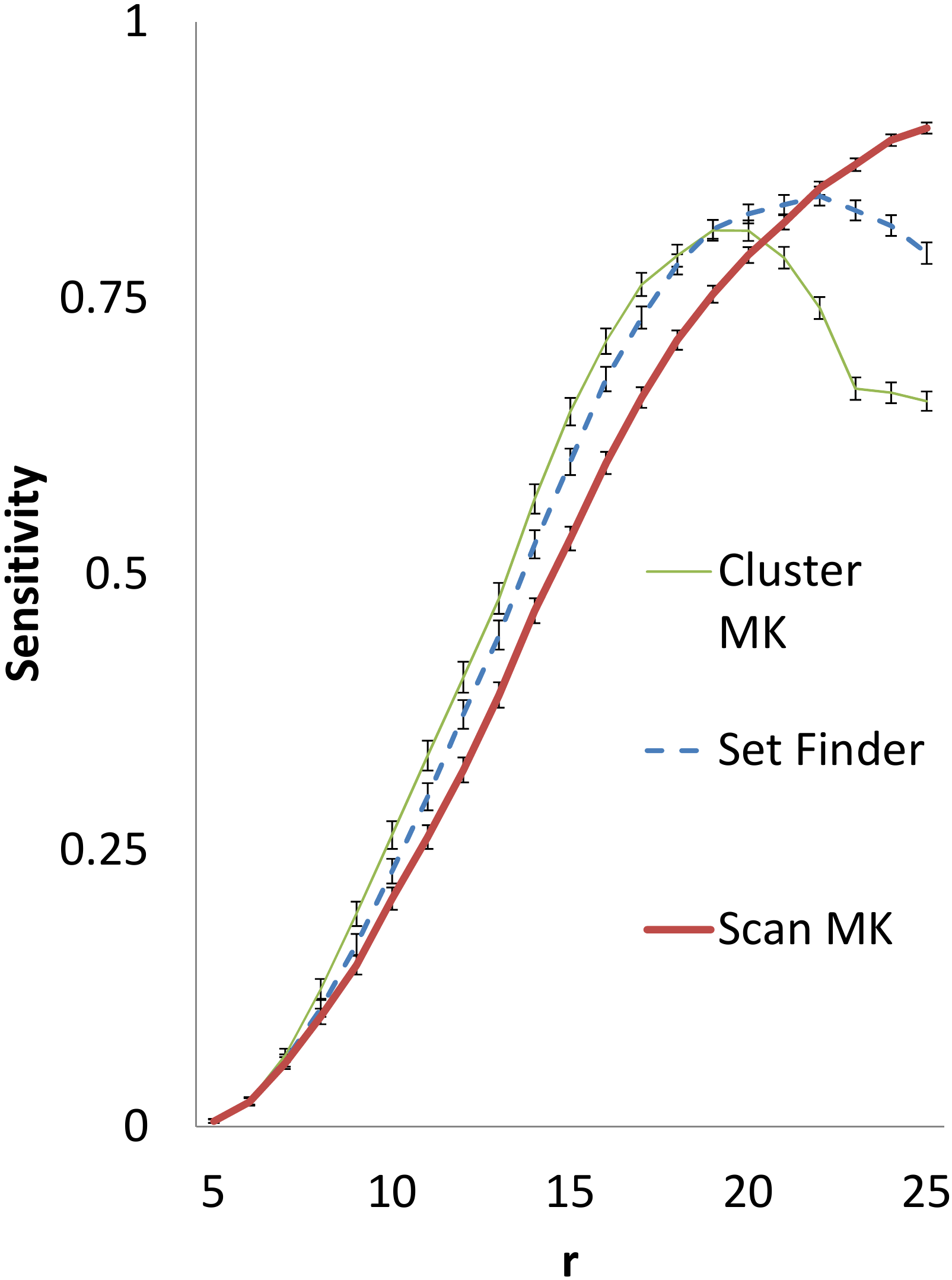}
\includegraphics[width=1.6in,height=2in]{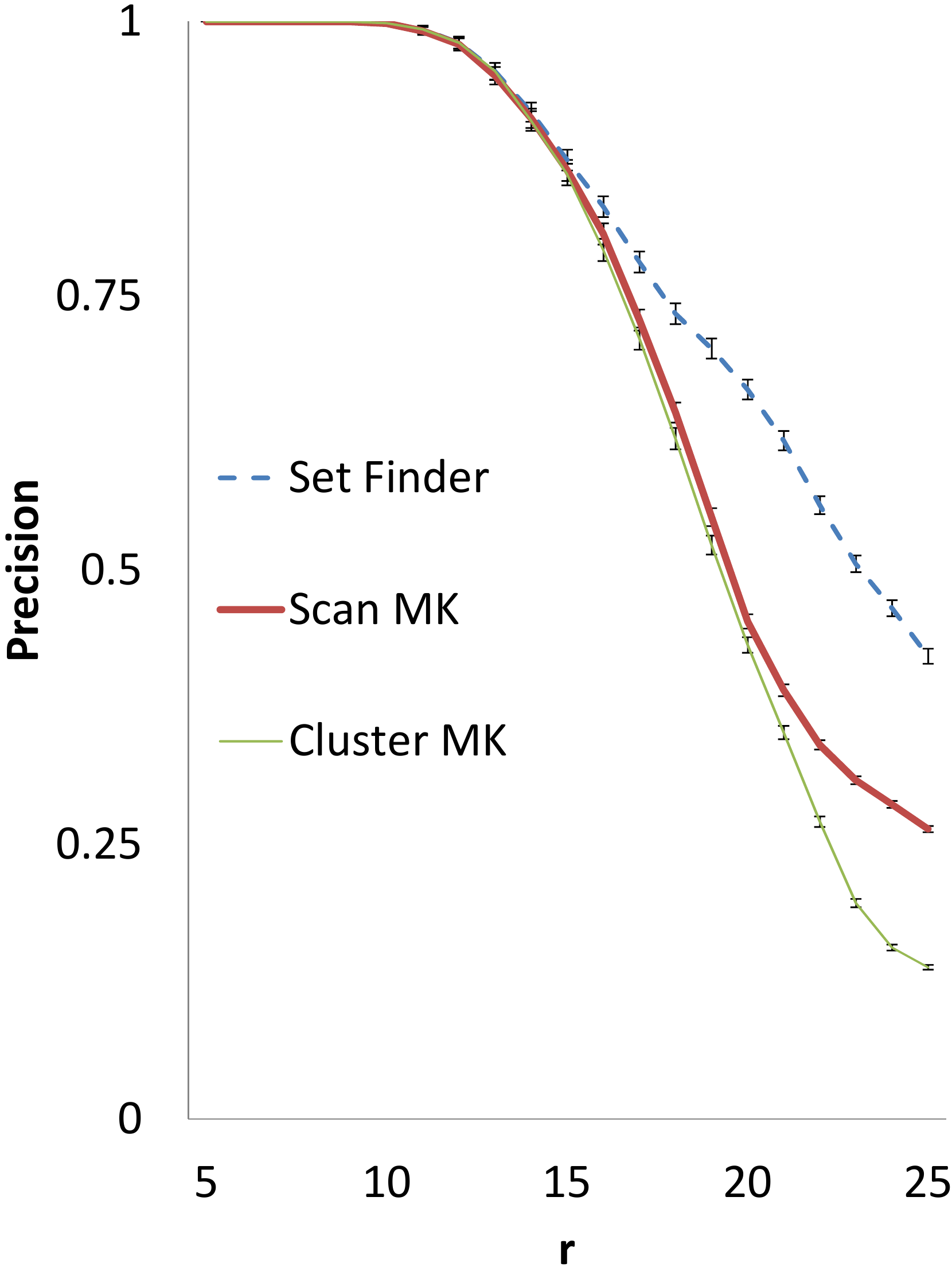}
\caption{The mean sensitivity (left) and precision (right), with standard error, over a random sample of 1000 instances of the single shape datasets for Scan MK, Cluster MK, and Set Finder, with values of $r$ in the range 5 to 25.}
\label{fig_5}
\end{figure}

For the experiments using a single shape, we tested the three algorithms over 1000 datasets containing three to five instances of the shape, using different values of $r$ in the range $r = 5 - 25$. Figure \ref{fig_5} shows the results of these experiments. The statistically significant differences are listed below:

\begin{itemize}
\item Cluster MK is significantly more sensitive than the other algorithms in the range $r = 9 - 15$, and less sensitive in the range $r = 21 - 25$. It is less precise than the others in the range $r = 21-25$.
\item Set Finder is significantly more precise than the other two in the range $r = 16 - 25$, and more sensitive than Scan MK in the range $r = 10 - 20$.
\item Scan MK is less sensitive than the other two algorithms in the range $r = 10 - 25$, and more sensitive in the range $r = 22 - 25$.
\end{itemize}

We conclude that Cluster MK and Set Finder are more accurate than Scan MK; Cluster MK is better for finding motifs, and Set Finder is better for avoiding false positives. Cluster MK has the disadvantage that it is very sensitive to the value of $r$, and its performance degrades quickly outside of the optimum range. Scan MK is significantly more sensitive at high values of $r$, but the concomitant loss of precision suggests that it will give many false positives in this range, which may be unsuitable for certain tasks.


\subsection{Problems with two motif sets.}

Finding multiple motif sets is more complex, not least because the algorithms must distinguish between the subsequences that belong to different sets (see Section \ref{perform}). For the single shape problem, we permitted multiple sets to be aggregated; for the two-shape problem, each set returned by the algorithm is assigned to at most one of the motif sets in the data. Hence, an output of a single set containing all of the instances of both shapes is rewarded only for finding one motif set, and punished for missing the other set and for false positives. Equally, if the algorithm finds all instances of both motif sets, but splits them into many different sets, it is punished accordingly.


\begin{figure}[!t]
	\centering
        \includegraphics[width=3.2in]{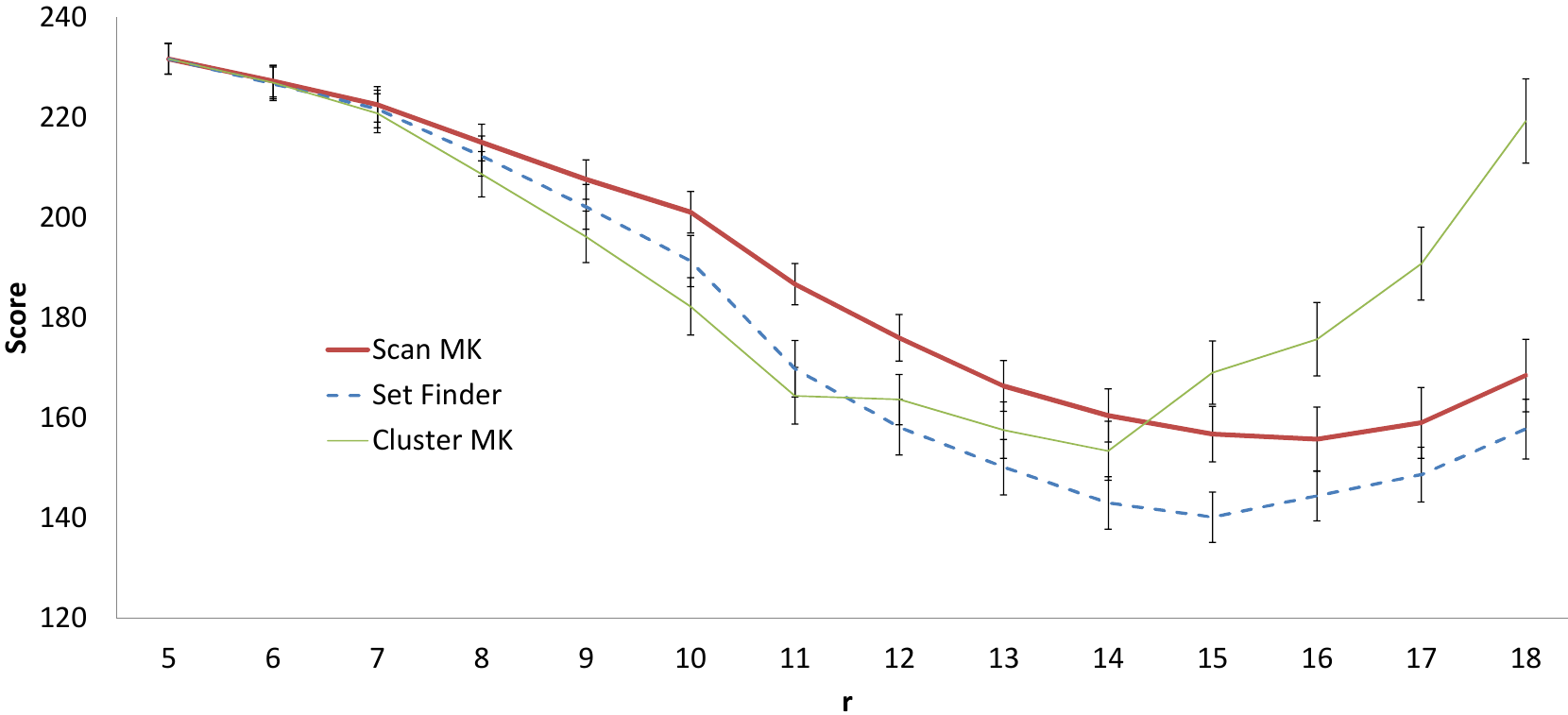}
	\caption{The mean difference score (as defined in Section~\ref{perform}) and standard error over a random sample of 100 instances of the two shape datasets for Scan MK, Cluster MK, and Set Finder with values of $r$ in the range 5 to 18. Lower scores indicate better performance. }
	\label{fig_7}
\end{figure}

The results of the two-shape experiment are shown in Figure \ref{fig_7}. The statistically significant differences are listed below:

\begin{itemize}
\item Set Finder is significantly better than Scan MK in the range $r = 11 - 15$ and significantly better than Cluster MK in the range $r = 15 - 18$.
\item Cluster MK is significantly better than Scan MK in the range $r = 9 - 12$; this is reversed in the range $r = 16 - 18$.
\item The best values of $r$ for all algorithms are in the range $r = 14 - 16$, when $r$ is approximately $\frac{n}{2}$.
\end{itemize}


Our results suggest that Set Finder is the most accurate algorithm when $r$ is approximately $\frac{n}{2}$, which we suggest is an appropriate value for noisy data. Once again, Cluster MK is very sensitive to the value of $r$, which is a weakness of the algorithm, as small differences in $r$ (which is difficult to estimate precisely), can cause the algorithm's performance to deteriorate.

\section{Electricity-usage Data}
\label{elecRes}

\begin{figure*}[htbp]
	\centering
        \includegraphics[width=14cm]{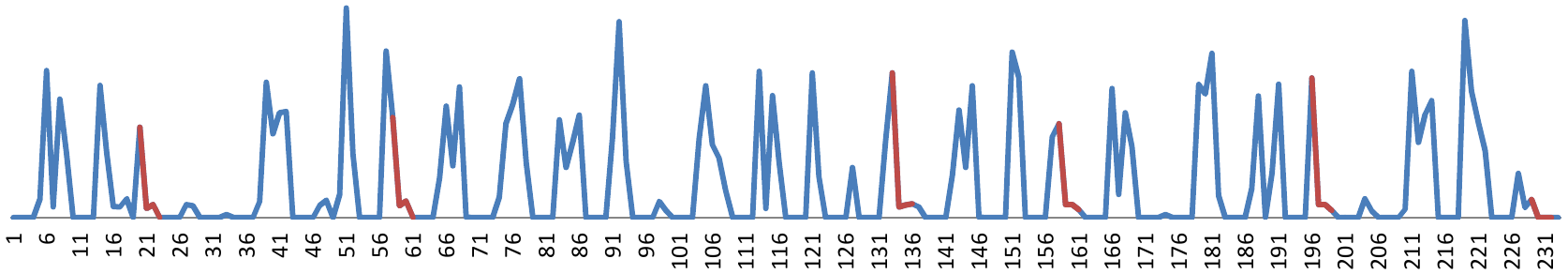}
	\caption{A household electricity-usage profile; the subsequences returned by the Set Finder algorithm as members of the two-motif set are highlighted in red, and represent the washing machine device.}
	\label{fig_8}
\end{figure*}

\begin{figure*}[htbp]
	\centering
        \includegraphics[width=14cm]{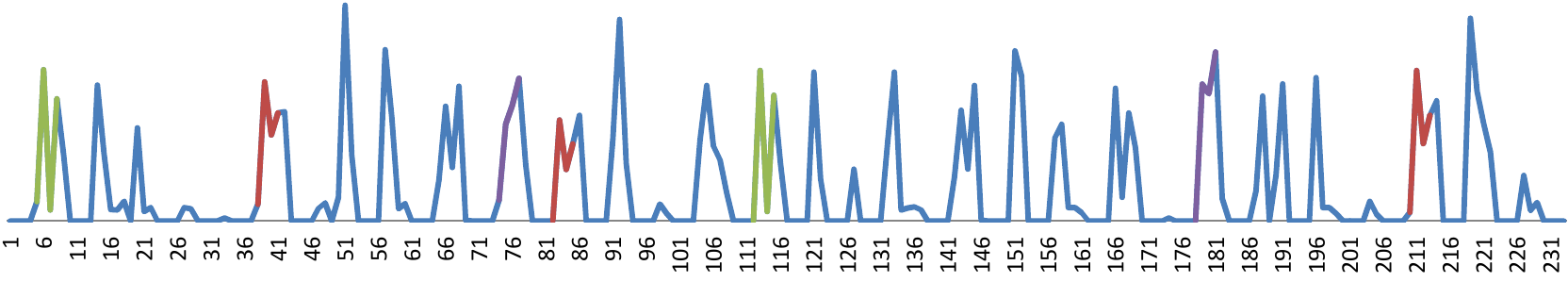}
	\caption{A household electricity-usage profile; the subsequences returned by the Scan MK algorithm are highlighted as follows: 4-motif set (red) and 5-motif set (green), are instances of the dishwasher device. The 6-motif set (purple) is two instances of the oven device.}
	\label{fig_9}
\end{figure*}

In this section, we present qualitative analysis on the performance of Set Finder and Scan MK on household electricity-usage data. We use a window size of 4, as this represents one hour. Our results show that analysis in terms of motif sets is likely to be fruitful for profiling device usage.

We first analyse the performance of the Set Finder algorithm on a usage profile. The usage profile contains usage instances of three devices: a dishwasher, a washing machine, and an oven. Using the values $r=0.5$ and $n=4$, the algorithm returns two sets of indexes. Unsurprisingly, the larger set contains all instances of four consecutive zeros, representing the instances of no device usage. More interestingly, the other set consists of indexes that closely resemble the usage profile of the washing machine. Figure \ref{fig_8} shows the usage profile with the discovered motif set highlighted in red. The 2-motif set found by the algorithm correctly identifies all instances of the washing machine in this usage profile, and no other devices.

The precision is 1 (no false positives), and for the washing machine device, sensitivity is also 1 (no false negatives). The sensitivity measured over all devices is 0.21; while this may appear to be fairly poor, it is better than some of the results obtained on the synthetic data.

The electricity-usage problem has an added level of complexity because the motif sets representing different devices contain subsequences of different lengths; this necessitates variation in the values of $n$ and $r$, and explains why the algorithm found one device perfectly, while missing the others. An appropriate approach for such data would involve producing output for many values of $n$ and $r$, and post-processing the discovered motif sets to find the set of devices in the data.

We turn now to the performance of the Scan MK algorithm. Again, we use fixed values of $n = 4$ and $r = 0.5$. As identified by the algorithm, the 1-motif set is largely 0 elements. As with the Set Finder algorithm, we disregard this set. We also disregard the 2-motif set and 3-motif set, as they contains very similar data that would be post-processed as belonging with the 1-motif set. The other sets of indexes returned by the algorithm are interpretable as follows. The algorithm has been reasonably successful at finding the dishwasher device, although post-processing would be required to combine the 4-motif set (highlighted in red on Figure \ref{fig_9}) and the 5-motif set (highlighted on Figure \ref{fig_9} in green). The precision of the combined set is 1 (no false positives); the sensitivity for the dishwasher is 0.56. The overall sensitivity is 0.24; this value includes the two oven devices identified as the 6-motif set (highlighted in purple in Figure \ref{fig_9}). It should be noted that the device-specific sensitivity of Scan MK was lower than that of Set Finder, even though Scan MK benefited from generous post-processing.


\section{Conclusion}
\label{conclusions}
Finding motif sets in time series is essentially a form of clustering, and it is necessary to define a heuristic search technique to find motif sets, as the problem is NP-complete. We have proposed and compared three such algorithms for motif-set discovery: Scan MK, Cluster MK, and Set Finder. Extensive experimentation shows that Set Finder is significantly more accurate than Scan MK on synthetic data containing one and two shapes, for the values of the range parameter $r$ for which the algorithms perform best. Cluster MK is competitive providing that appropriate values of $r$ are used; however, it is very sensitive to the value of $r$.

Finding motif sets is applicable to problems in a wide range of domains, including medicine, image processing, and robotics. We have extended our experiments to investigate the problem of profiling device usage from household electricity-consumption data. We found the motif set approach showed promise for identifying specific devices from data; we can reasonably expect to improve this performance dramatically on less aggregated data, and by using varying values of $n$ and $r$ followed by post-processing.


%
\ifCLASSOPTIONcompsoc
\else
\fi


\ifCLASSOPTIONcaptionsoff
  \newpage
\fi

\end{document}